# Detection of Microcalcification in Mammograms Using Wavelet Transform and Fuzzy Shell Clustering

T.Balakumaran
Department of Electronics and Communication Engineering
Velalar College of Engineering and Technology, Erode, TamilNadu, India.

Dr.ILA.Vennila
Department of Electrical and Electronics Engineering
PSG College of Technology, Coimbatore, TamilNadu, India

C.Gowri Shankar
Department of Electrical and Electronics Engineering
Velalar College of Engineering and Technology, Erode, TamilNadu, India.

*Abstract*— Microcalcifications in mammogram have been mainly targeted as a reliable earliest sign of breast cancer and their early detection is vital to improve its prognosis. Since their size is very small and may be easily overlooked by the examining radiologist, computer-based detection output can assist the radiologist to improve the diagnostic accuracy. In this paper, we have proposed an algorithm for detecting microcalcification in mammogram. The proposed microcalcification detection algorithm involves mammogram quality enhancement using multirresolution analysis based on the dyadic wavelet transform and microcalcification detection by fuzzy shell clustering. It may be possible to detect nodular components such as microcalcification accurately by introducing shape information. The effectiveness of the proposed algorithm for microcalcification detection is confirmed by experimental results.

I. INTRODUCTION

Breast cancer is the second leading cause of cancer deaths for women and is found as one in eight women in the United States. It is a disease in which cells in the tissues of the breast become abnormal and divide without order or control. These abnormal cells form too much tissue and become a tumor. According to WHO report, nearly two million women are diagnosed with breast cancer every year worldwide. The disease can be treated if discovered so early enough. The effective detection of breast cancer in earlier stage increases the survival rate. The appropriate method for early detection of pre cancerous symptoms is screening mammography, which has to be conducted as a regular test for women.

Calcification clusters are an early sign of breast Cancer. Microcalcifications are quite very small bits of calcium deposits present inside the soft breast tissue. It shows up in clusters or in patterns (like circles or lines) associated with extra cell activity in breast region.

Microcalcifications appear in mammogram image as small localized granular points with high brightness. It is not easy to detect by naked eye because of its miniaturized dimension. However about 10%-40% of Microcalcification clusters are missed by radiologists due to its small size and nonpalpable [1], [2]. To avoid these problems, a New CAD (computer Aided diagnosis) system has to be developed to improve the diagnostic rate. By incorporating the expert knowledge of radiologists, the CAD system can be made to provide a clear insight about the disease and saves the society from breast cancer.

Many researchers have proposed the algorithms for Microcalcification detection based on discrete wavelet transform, which is a powerful tool for analyzing the image hierarchically on the basis of scale. Some researchers have developed a CAD system using fuzzy clustering, artificial neural network and genetic algorithm. R. N. Strickland *et al.* [3], H.Yoshida *et al.* [4] used a wavelet transform to detect microcalcification and the fuzzy logic was tried by N.Pandey *et al.* [5]. Anne Strauss et al. [6] presented an identification scheme based on watershed Processing. Valverde *et al.* [7] used a deformable-based model for Microcalcification detection. Some of these studies detect approximately 70% to 80% of correct calcification. Objective of CAD system is to reduce the false positives and consistency of radiologists in image interpretation [8]. Naturally Microcalcifications are nodular in structure, other tissue such as mammary ducts blood vessels are linear in structure [9]. The Fuzzy shell clustering algorithm (FSC) is best in identifying circular objects present in an image [10]. In the proposed method, wavelet has been combined with fuzzy shell clustering (FSC) algorithm in order to mark the Microcalcification region

The rest of this paper is organized as follows. Section II presents microcalcification enhancement by using dyadic wavelet transform; the detection part using Fuzzy shell clustering in Section III, Results obtained on execution of algorithm are presented in section IV and Conclusion as last section.

## II. MICROCALCIFICATION ENHANCEMENT

The fundamental operation needed to assist microcalcification detection in mammogram is image contrast enhancement. In many image-processing applications, the grayscale histogram equalization (GHE) is one of the simplest and effective techniques for contrast enhancement. Histogram equalization improves contrast of the microcalcification in mammogram image by reassigning the intensity values of pixels based on the image histogram. But Histogram equalization technique doesn't enhance the microcalcification region because it modifies the intensity of the image with some unpleasant visual artifacts.

Microcalcifications appear as subtle and bright spots, whose size varies from 0.3mm to 1mm in the mammogram image. It is not easy to enhance the microcalcification regions since surrounding dense breast tissue makes the abnormality areas almost invisible. Microcalcifications are high frequency in nature. So it can be extracted by using high pass filtering. But conventional enhancement technique like unsharp masking, homomorphic filters and high boost filtering tends to change the characteristics of microcalcification. To overcome these limitations microcalcification regions can be enhanced by dyadic wavelet transform without modifying characteristics of microcalcification.

Wavelet analysis permits the decomposition of image at different levels of resolution. In Fig. 1, the filter bank structure of the two-dimensional wavelet transform is shown from level j to level j+1, which generates four sub-images at level j+1. $S_j$ be original image, the approximation sub-image $S_{j+1}$ is obtained by applying the vertical low-pass filter followed by horizontal low-pass filter to $S_j$. The sub-image $S_{j+1}^{LH}$ is obtained by applying the vertical low-pass filter followed by the horizontal high-pass filter. The sub-image $S_{j+1}^{HL}$ is obtained by applying the vertical high-pass filter followed by horizontal lowpass filter. Finally, the response $S_{j+1}^{HH}$ is obtained by applying the vertical and horizontal high-pass filters successively [9]. The downsampling by a factor 2 is introduced after each level of filtering. The same procedure is repeated for each level of approximation coefficients till $S_{j+n}$ is achieved.

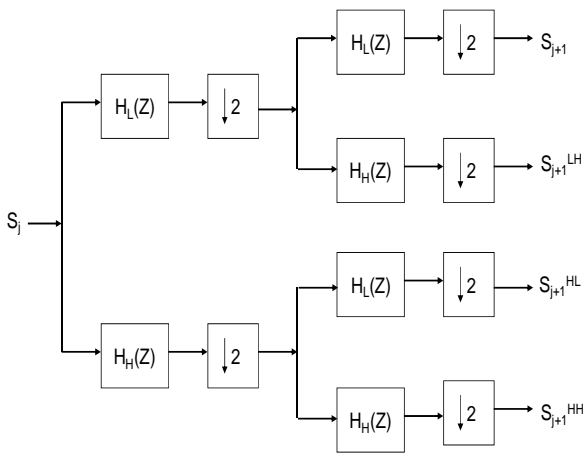

Fig. 1 Two dimensional dyadic wavelet structure

The digitized mammogram incorporated with a size of 1024 x 1024 pixel was taken from Digital Database for screening mammogram(DDSM). Mammogram image was decomposed upto 10 levels by applying dyadic wavelet transform with a decimation factor 2. The original 1024x1024 grayscale digital mammogram image was decomposed to 10 levels by applying Daubechies4 wavelet transform. Finally the lowest approximation image $S_{10}$ is of single pixel width. Since microcalcification appears as high frequency behavior in mammogram, the enhancement is achieved by setting the value of $S_{10}$ as zero.

The detail coefficients are enhanced by as per the expression number (1)

$$S_j^D(x,y) = \begin{cases} S_j^D(x,y) & \text{if } |S_j^D(x,y)| < T_j \\ g * S_j^D(x,y) & \text{if } |S_j^D(x,y)| \geq T_j \end{cases} \quad (1)$$

where, x and y are spatial coordinates, D represents all horizontal, vertical and diagonal subbands. $T_j$ be a non negative threshold obtained by taking standard deviation of respective subimage. The best visual quality of microcalcification is obtained while the gain(g) is set as 1.2. The reconstruction of weighted higher frequency subbands provides better visibility of microcalcification region than the other breast regions.

## III. MICROCALCIFICATION DETECTION

The CAD system was developed to assist the radiologist in detecting the suspicious areas on the basis of certain feature extracted from the images. After enhancing the microcalcification regions, next step in the journey of climbing the aim is the extraction of microcalcification through Soft computing tools. Microcalcification appears in mammogram as nodular points with higher brightness, localized or broadly diffused along the breast tissue, whereas normal tissues such as blood vessels are linear in structure. So detecting the nodular structure in image is a key in detecting the microcalcification.

### A. Region of Interest (ROI) Identification

The first stage of microcalcification detection is ROI identification. The enhanced mammogram image is decomposed by undecimated wavelet transform (filter bank implementation without downsampling). The resulting horizontal detailed image or vertical detailed image is used to identify the region encircling the microcalcification clusters. Third and fourth order statistical parameters, skewness and kurtosis [11], are used to find the regions of microcalcification clusters.

An estimate of the skewness is given by

$$S_k = \frac{\sum_{i=1}^{N}(x_i - \tilde{m})^3}{(N-1)\sigma^3} \quad (2)$$

and the statistical parameter kurtosis holds the expression

$$K_u = \frac{\sum_{i=1}^{N}(x_i - \tilde{m})^4}{(N-1)\sigma^4} - 3 \quad (3)$$

where $x_i$ is the input data over N observations, $\tilde{m}$ is the ensemble average of $x_i$ and $\sigma$ with its standard deviation. The third and fourth order statistical estimates were calculated for every overlapping 32x32 square regions of horizontal bandpass image or vertical bandpass image. The area having skewness value greater than 0.2 and kurtosis value greater than 4 is marked as a region of interest (ROI).

### B. Fuzzy Shell clustering

Extraction of features is the key process in the development of CAD system. Fuzzy shell clustering (FCS) algorithm is used to perceive nodular structure from the ROI. It extracts the clusters with spherical symmetry of a given data set $X=\{x_1,x_2,\ldots x_n\}$ with the prototype cluster centers $V=\{v_1,v_2,\ldots v_c\}$. Here c denotes the number of clusters formed during processing. Each data point $x_k$ has a degree of membership $u_{ik}$ to the i$^{th}$ cluster. The data point $x_k$ is assigned to a cluster if the below given weighted objective function is minimum.

$$J = \sum_{k=1}^{n}\sum_{i=1}^{c} u_{ik}^m D_{ik}^2 + \sum_{i=1}^{c} W_i \sum_{k=1}^{n}(1-u_{ik})^m \quad (4)$$

Where $D_{ik}$ is distance from data point $x_k$ to cluster center $v_i$ and $W_i$ be the 3-db width of the i$^{th}$ cluster. The weighting exponent $m \in (1,\infty)$ is the fuzziness controlling parameter with m as a real number greater than 1, it was chosen as 2 for better segmentation. Microcalcification appears as nodular (circular) in structure, the FCS algorithm identifies nodules described by their center $v_i$ and radius $r_i$. The geometric distance function of clustering is

$$D_{ik} = |\|x_k - v_i\| - r_i| \quad (5)$$

The necessary condition on membership leads to following equation used to update $u_{ik}$ for the minimization of objective function

$$U_{ik} = \frac{1}{\sum_{i=1}^{c}(D_{ik}/W_i)^{\frac{2}{m-1}}} \quad (6)$$

The following equation is used to speed up the algorithm

$$\|u^{j+1} - u^j\| < \varepsilon \quad (7)$$

$\varepsilon$ is minimum threshold value, which is used to minimize the number of iteration.

### C. Nodular extraction

After identifying ROI, the next step is to detect edges of the enhanced image. Edges are detected by applying first order derivatives. First order derivative is implemented by gradient operator ($G_x$ & $G_y$). The edges are detected by computing the gradient of each pixel in the enhanced image in x direction and y direction. The gradient of f(x,y) is

$$\nabla f = |G_x| + |G_y| \quad (8)$$

By applying first order derivative, Enhanced edges were detected. These enhanced edge pixels are connected in sets to form group. The Groups containing less than 5 pixels are discarded. Fuzzy Shell clustering algorithm was applied to only edge point, which belongs to ROI. FCS was applied several times with c= 1 and $W_i$ = 9. As a result, different circles with different radius were obtained. Final goal is to extract only circular (nodular) region, whose radius is equal to microcalcification radius.

After obtaining different circular regions, microcalcification detection is performed to distinguish between valid microcalcification region and invalid one. The two validity measurement parameters are Cluster density and Relative Shell Thickness [12]. These two parameters are used to identify microcalcification regions.

$$C_{di} = \frac{\sum_{k=1}^{n} u_{ik}}{2\pi r_i} \quad (9)$$

Numerator denotes sum of membership function on most characteristic points ($|u_{ik}| > 1/2$) and denominator denotes area of the circular region.

Relative shell Thickness is defined by

$$RST_i = \frac{\sum_{k=1}^{n} u_{ik}^m D_{ik}^2}{r_i \sum_{k=1}^{n} u_{ik}^m} \quad (10)$$

Where $r_i$ is radius of circular region. Correct microcalcification was detected according to the rule

If ($C_{di} > 1.15$ && $RST_i < 0.2$)
    Microcalcification nodular
End

The above thresholds were found by different experiments. Microcalcification was successfully detected by our proposed method.

## IV. RESULTS

To test the proposed method, experiments were performed on the set of mammogram image with different size and features which were obtained from DDSM database. In DDSM database, the size of each pixel is 0.0435mm and gray level depth is 12 bits.

Fig.2(a) shows a low contrast mammogram image of size 1024x1024. Fig.2(b) shows enhanced image where one can clearly visible microcalcification than the original image. Fig.2(c) shows Regions of Interest (ROI) was calculated by high order correlation parameters such as Skewness and kurtosis. Fig.2(d) shows edge map of enhanced image by applying first order derivatives. Fig.2(e) shows edge map after removal of small unwanted groups and Fig.2(f) shows the microcalcification region after applying FCS. Here FCS was applied several times to each edge point which is belonging to ROI.

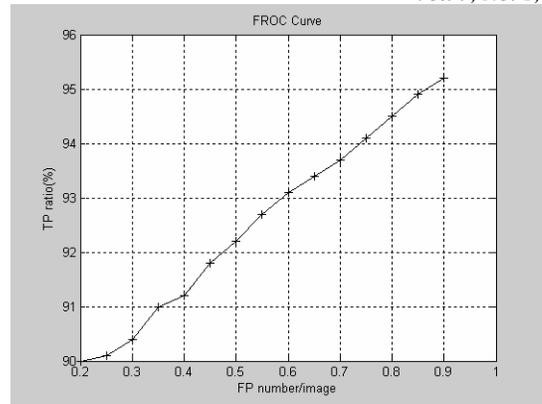

Fig. 3 FROC Curve of Microcalcification Detection

Free-response receiver operating characteristic (FROC) curve is used to evaluate the performance of microcalcification detection [13]. An FROC plot is achieved by calculating of the true-positive detection ratio (TP) with average number of false positives (FPs) per image. True positive detection ratio here refers that how many true abnormalities (microcalcifications) are correctly detected by computerized scheme and false positive per image refers to how many true abnormalities are missed. We have tested 112 images for FROC analysis. Our proposed method was achieved 95.23% TP ratio with 0.9 FP number/image.

## V. CONCLUSION

In this paper, an algorithm for image enhancement based on wavelet transform and microcalcification detection using a Fuzzy shell clustering were proposed. Original Mammogram image is decomposed by applying dyadic wavelet transform. Enhancement is performed by setting lowest frequency subband of decomposition to zero. Then image is reconstructed from the weighted detailed subbands, the visibility of the resultant's image is greatly improved. Experimental results were confirmed that microcalcification can be accurately detected by fuzzy shell clustering algorithm.

We proposed an algorithm for Microcalcification detection by introducing shape information. FCS algorithm can detect microcalcification if it's nodular in structure. We have tested112 images from our database. In these images, 95% of microcalcifications were detected correctly and 5% of microcalcifications were failed to detect because of they are not nodular in structure. Thus the research is still being performed to find out the better way to detect microcalcification without nodular structure.

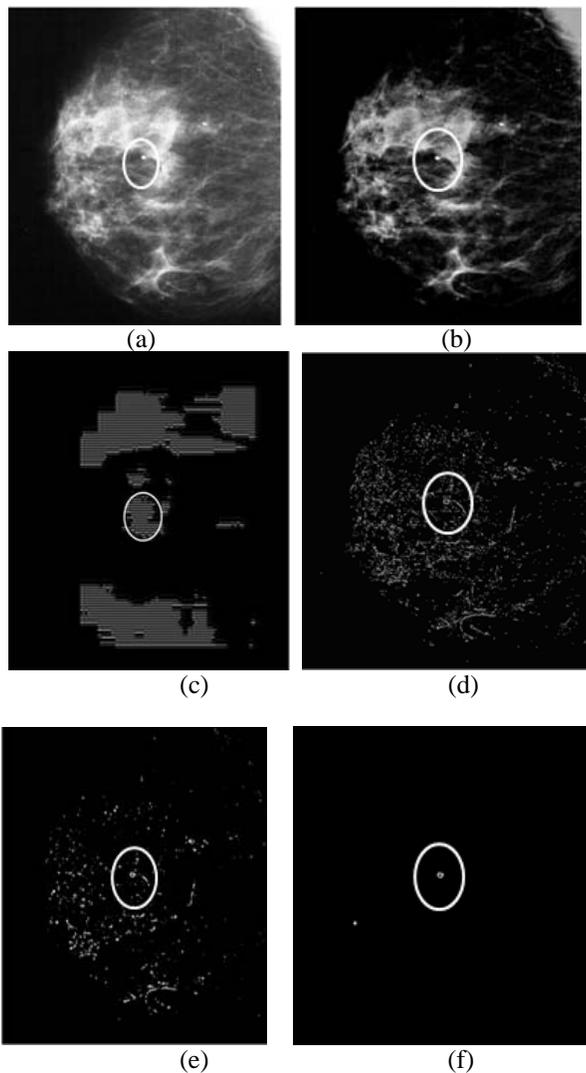

Fig. 2 Steps of microcalcification detection: (a) Original Mammogram Image (b) Enhanced image by wavelet transform (c) Region of Interest (ROI)identified by using Skewness & kurtosis (d) Edge map of enhanced image (e) Edge map after removal of unwanted groups (f) Microcalcification Detection by FCS algorithm

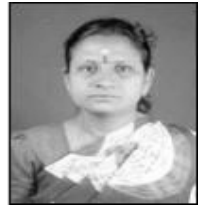

Dr.ILA.Vennila received the B.E Degree in Electronics and Communication Engineering from Madras University, Chennai in 1985 and ME Degree in Communication System from Anna University, Chennai in 1989. She obtained Ph.D. Degree in Digital Signal Processing from PSG Tech, Coimbatore in 2006. Currently she is working as Assistant Professor in EEE Department, PSG Tech and her experience started from 1989; she published about 35 Research Articles in National, International Conferences National and International journals. Her area of interests includes Digital Signal Processing, Medical Image processing, Genetic Algorithm and fuzzy logic.

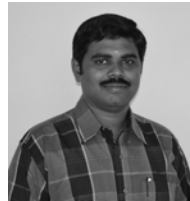

Mr.C.Gowri Shankar received the B.E. Electrical and Electronics Engineering from Periyar University in 2003 and M.E Applied Electronics from Anna University, Chennai in 2005. Since 2006, he has been a Ph.D. candidate in the same university. His research interests are Multirate Signal Processing, Computer Vision, Medical Image Processing, and Pattern Recognition. Currently, he is working in Dept of Electrical and Electronics Engineering, Velalar College of Engineering and Technology, Erode.

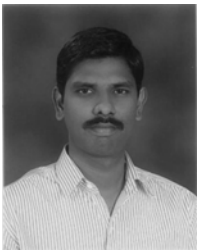

T.Balakumaran received the Bachelors degree in Electronics and Communication Engineering from Bharathiyar University, Coimbatore in 2003 and the Master degree in Applied Electronics from Anna University, Chennai in 2005. Since then, he is working as a Lecturer in Velalar College of Engineering and Technology (Tamilnadu), India. Presently he is a Part time (external) Research Scholar in the Department of Electrical Engineering at Anna University, Coimbatore (India). His fields of interests include Image Processing, Medical Electronics and Neural Networks.